\documentclass[letterpaper, 10 pt, conference]{IEEEtran}  
\IEEEoverridecommandlockouts                              
\usepackage{tabularx}
\usepackage{graphicx} % for pdf, bitmapped graphics files
\usepackage{times} % assumes new font selection scheme installed
\usepackage{amsmath} % assumes amsmath package installed
\usepackage{amssymb} % assumes amsmath package installed
\usepackage[linesnumbered,lined,boxed,commentsnumbered]{algorithm2e}
\usepackage{float}
\usepackage[noend]{algpseudocode}
\usepackage{multirow,color}
\usepackage{cite}
\usepackage{algpseudocode}
\usepackage{varwidth}
\usepackage{subfig}
\usepackage{booktabs}
\usepackage{gensymb}
\usepackage[hyphens]{url}
\usepackage[export]{adjustbox}
\usepackage[font=footnotesize]{caption}

\usepackage[inline]{enumitem} % For inline lists
\usepackage{graphicx,tipa}% http://ctan.org/pkg/{graphicx,tipa}
\usepackage{stfloats}

\usepackage{color} 
\usepackage{siunitx}
\usepackage[hidelinks]{hyperref}

\RestyleAlgo{ruled}
\usepackage{makecell}

\usepackage{verbatim} %In the preamble 

\makeatletter
\def\BState{\State\hskip-\ALG@thistlm}
\makeatother

\title{\LARGE \bf 
{Using Robotics to Improve Transcatheter Edge-to-Edge Repair \\ of the Mitral Valve}}
\author{Léa Pistorius$^{1}$, Namrata U. Nayar$^{1}$, Phillip Tran$^{1}$, Sammy Elmariah$^{2}$ and Pierre E. Dupont$^{1}$
\thanks{This work was supported by the NIH under grant R01HL124020.}
\thanks{$^{1}$ L. Pistorius, N. Nayar, P. Tran and P. Dupont are with the Department of Cardiac Surgery, Boston Children’s Hospital, Harvard Medical School, Boston, MA, USA.}
\thanks{$^{2}$S. Elmariah is with the Department of Cardiology, University of California San Franciso School of Medicine, San Francisco, CA, USA.}
\thanks{$\dagger$Pierre. E. Dupont is the corresponding author (pierre.dupont@childrens.harvard.edu).}
}

\begin{document}
\maketitle
\pagestyle{plain}
\begin{abstract}
\par{\;}{Transcatheter valve repair presents significant challenges due to the mechanical limitations and steep learning curve associated with manual catheter systems. This paper investigates the use of robotics to facilitate transcatheter procedures in the context of mitral valve edge-to-edge repair. The complex handle-based control of a clinical repair device is replaced by intuitive robotic joint-based control via a game controller. Manual versus robotic performance is analyzed by decomposing the overall device delivery task into motion-specific steps and comparing capabilities on a step-by-step basis in a phantom model of the heart and vasculature. Metrics include procedure duration and clip placement accuracy. Results demonstrate that the robotic system can reduce procedural time and motion errors while also improving accuracy of clip placement. These findings suggest that robotic assistance can address key limitations of manual systems, offering a more reliable and user-friendly platform for complex transcatheter procedures.}
\end{abstract}
\section{Introduction}\label{sec:intro}
\begin{figure}
        \centering
        \includegraphics[width=0.49\textwidth]{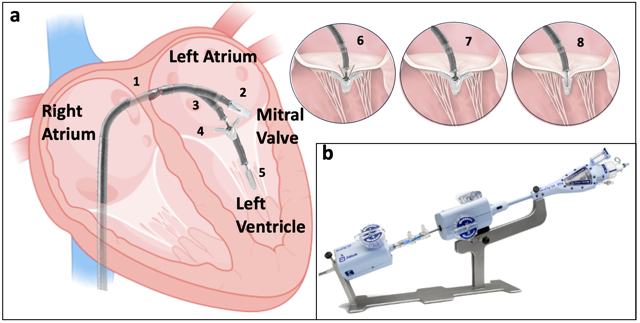}
        \caption{TEER system for correcting mitral valve regurgitation. (a) Delivery steps. (b) Sheath control handles mounted in stabilizer. Adapted from \cite{MitraClip}.}
        \label{steps}
        \end{figure}

Transcatheter valve repair procedures are complex to perform and involve substantial learning curves. For example, in transcatheter edge-to-edge repair (TEER) of mitral regurgitation (Fig. \ref{steps}), clinical experience demonstrates that operators improve significantly over their first 50 cases and their performance continues to improve out to their 200th case \cite{Chhatriwalla}.

A major component of mastering a transcatheter procedure is learning how to precisely control catheter positioning. These interventions are performed using delivery systems comprised of nested sets of telescoping steerable sheaths. Manual control of the catheter tip is achieved by sequentially rotating and translating each individual sheath, as well as by using the knobs on each handle to bend the sheath. To keep the other sheaths from moving while one is adjusted, all the handles are often locked in a stabilizer positioned above the patient's leg (Fig. \ref{steps}b). Inadvertent motions of the locked sheaths often occur which, if not detected and corrected for, result in high-risk situations and poor procedural results \cite{Ch26}.

While robotic control provides the potential to improve these procedures by enabling intuitive control of catheter tip motion, there are challenges. The kinematics of catheters are difficult to model and exhibit substantial hysteresis \cite{Yuan2025}. While sensors, such as electromagnetic (EM) trackers and fiber Bragg gratings can be used to close the kinematic control loop \cite{Qi2023}, they can be challenging and expensive to integrate. 

To investigate the potential for robotic improvement, several research groups have developed drive systems for clinical TEER devices. For example, \cite{DeMomi1} and \cite{XChen} construct robotic drive systems that actuate the device’s handles and knobs. To control catheter motion, \cite{DeMomi1} developed an open-loop kinematic model while \cite{XChen} created an interface in which three joysticks were used to control motions of the three handles. 

\begin{figure}[h!]
        \centering
        \includegraphics[width=0.49\textwidth]{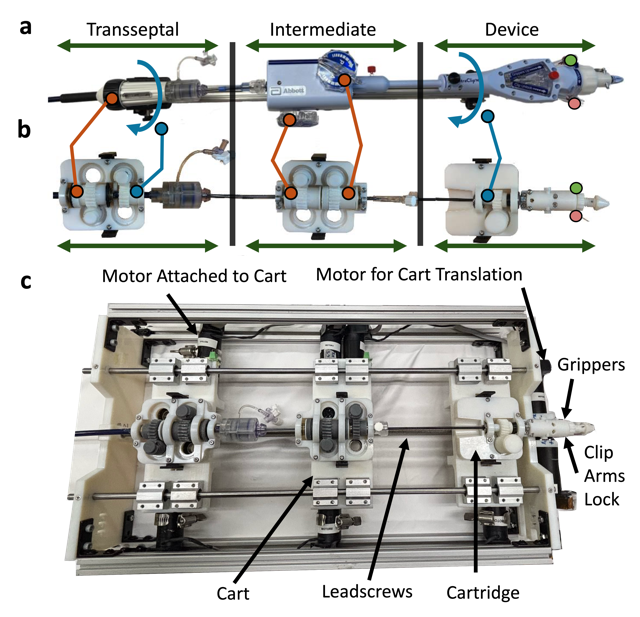}
        \caption{Conversion to robotic system. (a) Manual handles of TEER system. (b) Robotic cartridges. (c) Cartridges mounted in drive system.}
        \label{design_systems}
        \end{figure}

In this paper, a different approach is taken. Since manual handles are not designed for precision robotic control, they were removed and replaced with compact transmission cartridges that could interface with a universal catheter drive system (Fig. \ref{design_systems}). This approach also results in a smaller robot footprint. Furthermore, device delivery is decomposed into steps and joint-space control modes are developed for each step. In each control mode, intuitive motion control is enabled by mapping joystick motions from the game controller to the corresponding motions in the standard imaging view for that step. This approach eliminates the need for a complex kinematic model while significantly improving the precision and responsiveness of tip control. To demonstrate the effectiveness of this approach, we compare our model-free, but step-specific control strategy with manual device delivery in a phantom model.

The paper is organized as follows. In Section II, we decompose the delivery of a TEER device into 8 steps and, for each, identify the required motions and challenges. Section III describes the design of the robotic system. Experiments are described in Section IV and conclusions are presented in Section V.

\section{Stepwise Decomposition of TEER}\label{sec:analysis_TEER}

During normal mitral valve function, the two mitral leaflets press against each other, closing the valve and preventing abnormal backflow of blood from the left ventricle to the left atrium. However, during mitral regurgitation, the leaflets do not close properly, creating a jet of leaking blood. To correct this, one or more TEER devices are used to clip the leaflets together at the location of the jet. If performed correctly, a non-leaking double orifice valve is created with the leaflets opening and closing on each side of the clips. Successful repair depends on (1) placing the clips at the location of the jet, (2) aligning the axis of the clip with the axis of the valve, (3) orienting the clip arms perpendicular to the free edges of the leaflets and (4) ensuring that the clip arms completely capture the leaflets \cite{Sherif2017}.

The TEER delivery system used in this paper is comprised of three telescoping catheters possessing eight degrees of freedom (DOF), as shown in Fig. \ref{design_systems}. The degrees of freedom are as follows. The transseptal sheath can bend in one direction while also translating along and rotating about its axis. The intermediate sheath translates along its axis and can bend in two orthogonal directions corresponding to the medial-lateral and anterior-posterior planes. The device sheath can translate along and rotate about its axis, but does not bend. It possesses additional DOF for opening and closing the clip arms and the leaflet grippers (Fig. \ref{design_systems}c). Manipulation of the clip arms and grippers does not affect device positioning and are not motorized in the described implementation. The TEER system is typically introduced through the right femoral vein and navigated to the right atrium from where it is introduced into the left atrium through a puncture in the atrial septum. Device delivery starts with navigating the transseptal sheath across the atrial septum. Maneuvering and deploying the clip then involves eight distinct procedural steps, as  illustrated in Fig. \ref{steps} and detailed in Table \ref{tab:limitations} \cite{Ch26}. As indicated in the table, six of the eight steps pose challenges to the operator. Four of the six problematic steps relate to inadvertent joint motions (steps 2,4-6) while the challenge of step 3 is to produce the desired coordinated motion using iterative motions of four joints.

\begin{table}[t!]
    \centering
    \caption{Transcatheter Edge-to-Edge Repair Steps}
    \resizebox{\columnwidth}{!}{
\begin{tabular}
{|m{0.1\columnwidth}|m{0.3\columnwidth}|m{0.3\columnwidth}|m{0.3\columnwidth}|}
        \hline
        \textbf{Step (Mode)} & \textbf{Movements} & \textbf{Motions Required} & \textbf{Problems} \\
        \hline
        1 (1) & Advance TS into LA & TS translation & None \\
        \hline
        2 (2) & Advance Clip into LA \& direct toward annulus & IS translation, IS M/L bending, TS rotation& Inadvertent extension of DS \\
        \hline
        3 (3) & Position clip above regurgitant jet & IS translation, TS rotation, M/L IS bending, A/P IS bending (IS rotation) & Four different adjustments to produce desired motion  \\
        \hline
        4 (4) & Open clip arms \& rotate clip perpendicular to coaptation line& Clip arm open, DS rotation, DS translation& Clip rotation requires axial dithering; \& Clip roll displaces IS\\
        \hline
        5 (4) & Close clip arms \& advance clip through mitral valve& Clip arm close, DS translation (in)& Unwanted clip twisting due to stored torque \\
        \hline
        6 (4) & Open clip arms \& position leaflet segments on clip arms& Clip arm open, DS translation (out)& Unwanted clip twisting due to stored torque  \\
        \hline
        7 (4) & Capture leaflets & Lower grippers & Extent of leaflet capture not imaged \\
        \hline
        8 (4) & Clip close, lock, and release & Clip close, lock, and release, sheath retraction & None \\
        \hline
        \multicolumn{4}{|c|}{\textbf{Note:} Mode = Control mode, TS = transseptal sheath, IS = intermediate sheath, } \\
        \multicolumn{4}{|c|}{DS = device sheath, M/L = Medial/Lateral, A/P = Anterior/Posterior, } \\
        \hline
    \end{tabular}}
    \label{tab:limitations}
\end{table}

The overall clip delivery process proceeds as follows. Initially, the transseptal sheath is extended into the left atrium (step 1). Next, the intermediate sheath is extended and flexed to direct the clip down toward the mitral valve (step 2). This motion typically also involves rotating the transseptal sheath. While the device sheath is supposed to remain retracted, interaction forces and moments between the sheaths cause it to unintentionally extend up to 1cm, which could cause it to hit the wall of the left atrium. In step 3, the clip is positioned above the regurgitation jet such that the axis of the clip is parallel to the axis of the valve. While this is a five DOF positioning and alignment task, it can typically be accomplished using 3 DOF of the intermediate sheath and rotation of the transseptal sheath. Manually, these DOF must be sequentially adjusted to obtain the desired configuration. 

In step 4, the clip arms are extended for better visualization of clip roll angle and the device sheath is rotated (rolled) to align the clips orthogonal to the closure line of the leaflets. To release torsional friction between the device and intermediate sheaths, an axial dithering motion is applied manually to the device sheath during rotation. Despite the dither, torsional friction can cause the intermediate sheath to deflect, moving the clip position away from the jet. In this situation, it can be necessary to return to step 3 to correct clip positioning. 

% \begin{figure}[h!]
%         \centering
%         \includegraphics[width=0.49\textwidth]{figures/steps_adjustments.png}
%         \caption{\textbf{Sequential Ordering of the Procedural Steps with Adjustments.}}
%         \label{steps_flowchart}
%         \end{figure}

Once the clip arms are oriented correctly, the arms are closed and the clip is advanced into the left ventricle (step 5). During translation, stored torsional friction between the intermediate and device sheaths is often released resulting in unintended rotation of the clip. Clinically, if twisting is small, the operator can rotate the clip inside the ventricle. For larger amounts of twist, the clip must be retracted into the left atrium and steps 4 and 5 are repeated. Once clip roll angle is correct, step 6 is performed in which the clip arms are opened and the device sheath is retracted so as to capture the leaflets in the clip arms. During retraction, the operator needs to monitor for additional unintentional device roll and correct for it as needed.

In step 7, the leaflet grippers are used to grasp the leaflets between the grippers and clip arms. The challenge of this step is that the ultrasound imaging used to guide these steps does not reveal if the leaflets are fully captured between the grippers and clip arms. If ultrasound imaging shows that the jet has been substantially reduced or eliminated, the final step 8 involves closing and locking the clip arms and releasing the device. If necessary, the cardiologist can add multiple clips until there is an adequate reduction in mitral regurgitation\cite{Whitlow, Bertolini}.

% Mitral valve leaflets close to form a coaptation line leading to the creation of three regions: A1P1, A2P2, and A3P3 (Fig. \ref{procedure_success}A). Procedural success is considered when the clip is correctly placed on the regurgitation jet of the mitral valve and orthogonal to the mitral valve leaflets. The clip must be oriented perpendicular to the coaptation line (Fig. \ref{procedure_success}B).

% \begin{figure}[h!]
%         \centering
%         \includegraphics[width=0.49\textwidth]{figures/procedure_success.png}
%         \caption{(A) Mitral Valve Components (B) Procedural Success and Clip Measurements.}
%         \label{procedure_success}
%         \end{figure}

\section{Robotic TEER System}\label{sec:RoboticMitraClipSystem}

To determine whether robotic control could address these delivery challenges, the handles of the TEER system (Fig. \ref{design_systems}a) were replaced by transmission cartridges that connect to a universal catheter drive system (Fig. \ref{design_systems}b). The drive system provides up to four motorized inputs per cartridge and can also translate each cartridge independently. Eight DOF are used in this application: three to translate, rotate, and bend the transseptal sheath; three to translate and bend the intermediate sheath in two planes, and two to translate and roll the device sheath. The details of cartridge design are described below.

\subsection{Conversion of Manual Handles to Robotic Cartridges}

For the device sheath, translation is driven by the movement of the cart on the robotic platform (shown with green arrows in Fig. \ref{design_systems}b). Rotation, represented by a blue arrow on Fig. \ref{design_systems}a and by a blue dot on Fig. \ref{design_systems}b, is achieved using a motor-driven worm gear and worm wheel, which rotates the device sheath via a clamp system attached to the handle. The clamp ensures the sheath rotates with the handle, minimizing hysteresis and improving precision. The clip actuation mechanism is modified from the manual system. Tension is maintained on the locking cords to control the clip arms and on the wires to manipulate the grippers. Spring buttons act as a mechanical lock, holding the wires and cords individually in place when the buttons are released. A knob on the distal part of the handle actuates the opening and closing of the clip arms, while a pin prevents the release of the clip. Withdrawing the pin from its groove allows the clip to be released by rotating the cylinder counterclockwise.

The intermediate sheath provides 3DOF (Fig. \ref{design_systems}a): translation (green arrow), medial-lateral bending, and anterior-posterior bending (orange dots). Translation is driven by the robotic platform's linear cart movement. To replace the anterior-posterior bending plane, rotation was added for sheath adjustments. Medial-lateral bending is achieved using one of the original tendon pairs, which are pulled by a motor-driven worm gear mechanism. The worm gear, engaging a worm wheel, rotates a barrel that linearly advances or retracts the tendons. The tendons are fixed inside the barrel to prevent rotation and are protected from damage. For rotation of the sheath (blue dot Fig. \ref{design_systems}b), a similar worm gear system is used. The motor actuates the gear, turning the worm wheel, which rotates the sheath. This mechanism ensures the sheath remains fixed during bending and only rotates when intended. Thrust bearings prevent hysteresis and friction between bending and rotating components. Large washers on the barrel reduce unwanted motion and limit motor strain on tendon movement. Additional bearings stabilize the cartridge and center the sheath. To maintain a leak-proof handle, the original catheter’s hemostatic valve is used, and the tendons are sealed in a separate compartment, preventing backflow into the sheath.

The transseptal sheath (Fig. \ref{design_systems}a) provides 3DOF: translation (green arrow), rotation (blue arrow), and bending (orange dot). The transseptal sheath handle was adapted for the robotic platform (Fig. \ref{design_systems}b) using a design similar to the intermediate robotic cartridge with worm gears for bending and rotation, and the robotic platform cart for translation. The hemostatic valve at the proximal end of the catheter prevents leaks. The tendons are housed in a separate channel, distinct from the liquid channels, ensuring a leak-proof system.

\subsection{Catheter Control}
\begin{figure}[h!]
        \centering
        \includegraphics[width=0.49\textwidth]{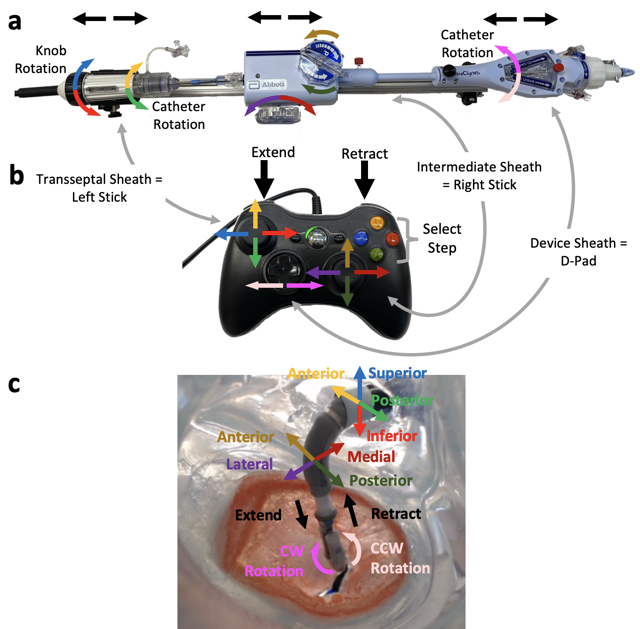}
        \caption{Catheter control. (a) Manual handle system, (b) Game controller, (c) Sheath motions from 3D en-face view of mitral valve.}
        \label{game_controller}
        \end{figure}

Our strategy is to enable intuitive and smooth catheter motion using a game controller that avoids the problems identified in Table I and without the need to implement and calibrate a kinematic model. Fig. \ref{game_controller} shows how the degrees of freedom of the three sheaths are mapped to the inputs of the game controller. With the clinician standing on the right side of the supine patient, the sheath handles, from left to right, control the transseptal sheath, intermediate sheath and device sheath (Fig. \ref{game_controller}a). To maintain this directionality, the transseptal and intermediate sheaths, which are responsible for most of the catheter motion, are assigned to the left and right joysticks, respectively, of the game controller (Fig. \ref{game_controller}b). 

While the joystick directions are mapped to individual sheath degrees of freedom, the mapping is defined such that the joystick directions intuitively correspond to sheath directions of motion when the catheters are imaged in the 3D en-face ultrasound view that is used to guide these procedures. As shown in Fig. \ref{game_controller}c, for the transseptal sheath, flexure produces predominantly vertical motion in the image view (superior / inferior motion) and is controlled using left and right motions of the left joystick. Rotation of the transseptal sheath predominantly produces motion across the valve from back to front (anterior / posterior) and is controlled by back to front motion of the left joystick.

Motion control of the intermediate sheath with the right joystick is similar except that in-plane flexure produces left to right motion across the valve (lateral / medial) and so is mapped to the left and right joystick directions. Sheath rotation still corresponds to anterior / posterior motion and is controlled by back to front motion of the joystick. Finally, insertion and roll of the device sheath are controlled using the D-pad of the game controller.
 
Sheath extension and retraction are controlled using the trigger buttons on the front of the controller. Four control modes, associated with specific steps of the delivery (see Table I), are selected using the X, Y, A and B buttons. The control mode determines which sheath translations are mapped to the trigger buttons. Furthermore, to avoid accidental input motions from the game controller, the modes only enable the degrees of freedom which are intended to be used during the step. A TEER device deployment consists of moving through these control modes to complete all the delivery steps. If it is necessary to repeat a step, the operator can return to the control mode associated with that step. 

The first mode is used for step 1, navigation of the transseptal sheath into the left atrium (Fig. \ref{steps}). For this mode, only the left joystick is active to steer the transseptal sheath and the trigger buttons advance and retract all three sheaths in unison. The second mode is used for step 2, advancing the clip into the left atrium and steering it down toward the mitral valve. In this mode, transseptal sheath translation and flexure are locked as are intermediate and device sheath rotation. The trigger buttons control simultaneous translation of the intermediate and device sheaths. Mode 3, for step 3, enables 3 DOF motion of the intermediate sheath and rotation of the transseptal sheath with the controller buttons commanding combined intermediate and device sheath translation. Mode 4 is used for steps 4-6 and allows only translation and roll of the device sheath along with rotation of the transseptal sheath. Device rotation using the D-pad buttons produces simultaneous automated translational dither motions (amplitude=2.5mm, frequency=2.2Hz) to reduce torsional windup. 

The game controller inputs are interpreted as velocity commands to the motors. To ensure safety while providing speeds comparable to manual operation, maximum motor speeds were constrained to produce maximum sheath speeds of $5.46^\circ/$sec in flexure, $14.56^\circ/$sec in roll and 6mm/sec in translation. 

\section{Experiments}\label{sec:Experiment}

\begin{figure}[h!]
        \centering
        \includegraphics[width=0.49\textwidth]{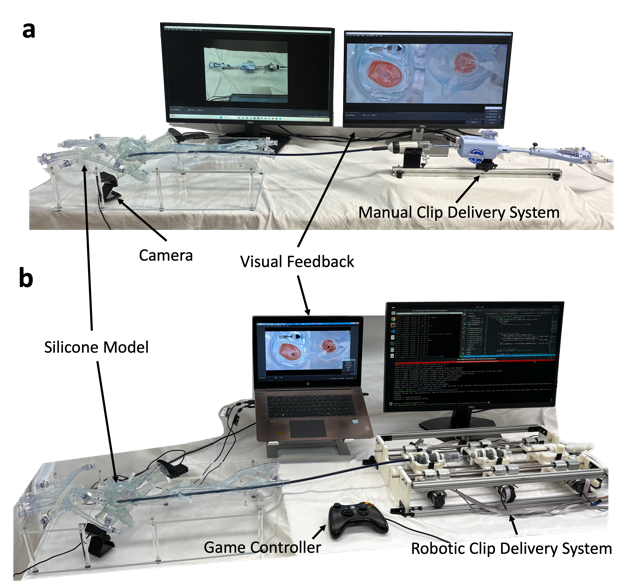}
        \caption{Experimental TEER system. (a) Manual system. (b) Robotic system.}
        \label{setup}
        \end{figure}

\begin{figure}[h!]
        \centering
        \includegraphics[width=0.49\textwidth]{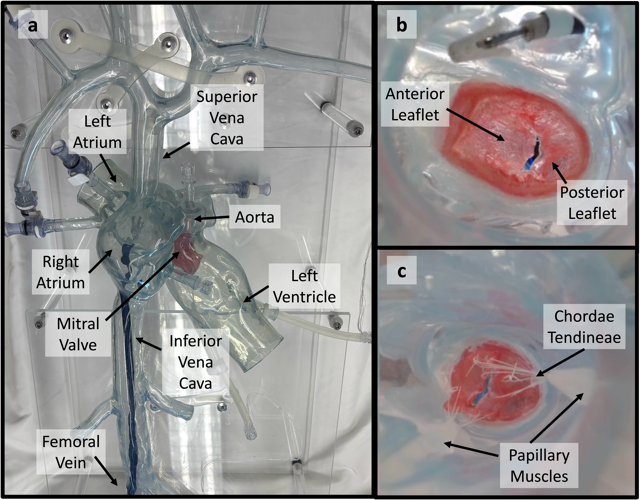}
        \caption{Mitral valve phantom model and camera views. (a) Phantom heart model. (b) En face camera view of mitral valve from left atrium. (c) En face camera view of mitral valve from left ventricle.}
        \label{heartmodel}
        \end{figure}

The TEER system used in the experiments was comprised of a transseptal sheath (Edwards PASCAL Precision System) combined with a Mitraclip TEER device (Abbott Mitraclip NTR). As shown in Figs. \ref{setup} and \ref{heartmodel}, experiments were performed in an anatomically accurate silicone model designed for evaluating transseptal delivery of devices to the mitral valve (United Biologics). To maintain anatomical orientation, a custom acrylic stand aligned the model’s vasculature and chambers to mimic patient positioning during mitral procedures. Catheters were inserted via the femoral vein and navigated through the inferior vena cava into the right atrium and then across the atrial septum into the left atrium. From there, the catheter flexes down to reach the mitral valve. 

The model incorporates two viewing ports that provide atrial and ventricular en face views of the mitral valve (Fig. \ref{heartmodel}b,c). Cameras (Logitech C920) positioned at the viewing ports stream real-time video to the operator (Fig. \ref{setup}). The video is also recorded (OBS Studio) for subsequent evaluation. An additional camera captured the motions of the operator’s arms and hands. 

Two sets of comparison experiments were performed. The first set investigated the motion problems identified in Table I to understand if the robotic implementation reduced or eliminated these problems.  The second set of experiments compared manual and robotic clip placement on each of the three segments of the mitral valve. Each set of experiments is described below. All experiments were performed by the same operator (LP).

\subsection{Comparison of Motion Problems Identified in Table I}

Three sets of experiments were performed to test if robotics could overcome the problems of steps 2-6 given in Table I. Each set involved 5 manual and 5 robotic trials. In the first set of experiments focused on Step 2, the manual and robotic catheters were steered down toward the annulus and inadvertent extension of the device sheath was measured. The second set of experiments considered Step 3 in which manual and robotic clip positioning over the valve were compared. Manual positioning involved sequential adjustment of handle motions for four degrees of freedom. Robotic positioning was accomplished using the game controller. While more practice might enable adjusting all four inputs simultaneously, the robotic trials were performed adjusting pairs of inputs sequentially. 

The third set of experiments evaluated the challenges of Steps 4-6. The clip, positioned above the valve with arms open and oriented parallel to the coaptation line, was rotated $90^{\circ}$ such that the arms were perpendicular to the coaptation line (step 4). The clip was then advanced down to the leaflets (similar to step 5) and then retracted to the original location above the valve (similar to step 6). At this point, any inadvertent rotation of the clip arms was measured. During the initial deliberate clip rotation of step 4, dithering motions were generated by hand in the manual experiments. In the robotic experiments, the controller provided an automatic axial oscillation whenever the user commanded clip rotation.

\begin{figure}[h!]
        \centering
        \includegraphics[width=0.49\textwidth]{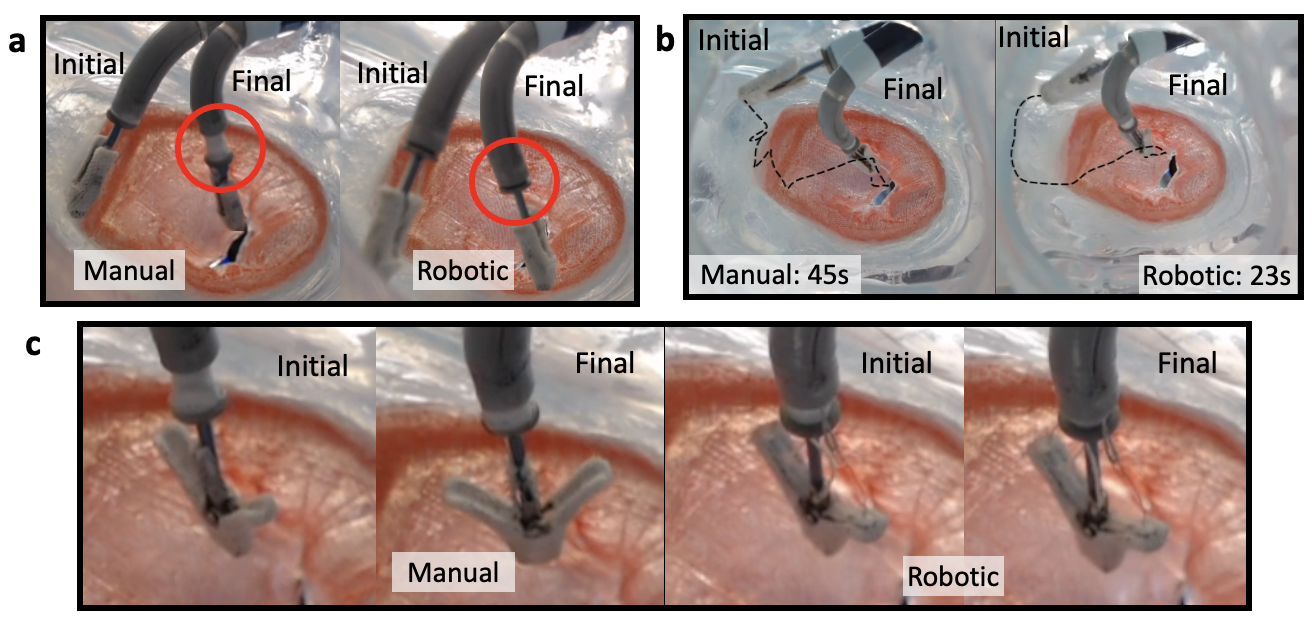}
        \caption{Robotic reduction of Table I delivery problems. (a) Step 2: Uncontrolled device sheath extension. (b) Step 3: Coordinated sheath motion. Clip tip paths appear as dashed curves. (c) Steps 4-6: Translation-induced twisting of device sheath.}
        \label{fig:results}
        \end{figure}

Example results for the first set of experiments, evaluating step 2 of TEER deployment, are shown in Fig. \ref{fig:results}a. As expected, manual flexure of the intermediate sheath to orient the clip toward the valve resulted in an average uncontrolled extension of the device sheath of $6\pm2$mm. In contrast, no device sheath extension was detected in the robotic system.

Fig. \ref{fig:results}b illustrates example results for producing the coordinated motion needed in step 3 to position and orient the clip above the regurgitant jet on the valve. The manual motion is comprised of 4 sequential motions involving 4 different degrees of freedom. Consequently, it moves the clip over a larger volume of the left atrium, increasing the risk of an accidental collision with the heart wall. In contrast, the robotic displacement moves smoothly toward the goal configuration. 

Fig. \ref{fig:results}c shows an example of the third set of experiments intended to measure inadvertent rotation of the clip arms during translation of the device sheath. As shown, the manual system produced uncontrolled rotations averaging $28\pm4^\circ$. Rotations in the robotic system were reduced to $6\pm7^\circ$.

\subsection{Comparison of Manual and Robotic Clip Delivery}

\begin{figure}[h!]
        \centering
        \includegraphics[width=0.27\textwidth]{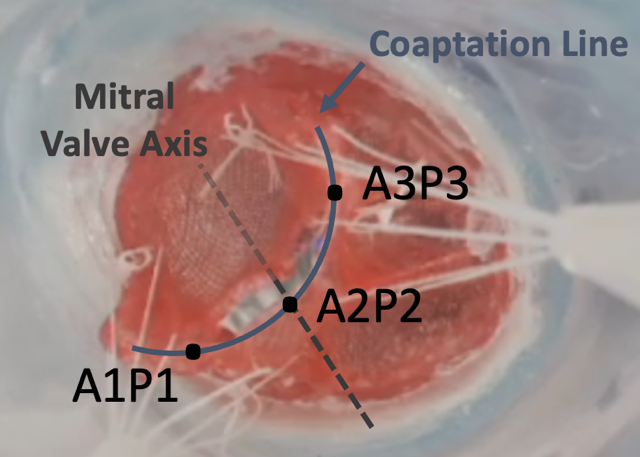}
        \caption{Clip placement targets. Targets comprised the centers of the leaflet segments with the clip oriented along the axis of the valve and with the arms orthogonal to the coaptation line.}
        \label{precision_metrics}
        \end{figure}

A single operator performed 30 manual trials followed by 30 robotic trials. The three leaflet segments, A1P1, A2P2 and A3P3 (Fig. \ref{precision_metrics}) were sequentially targeted in the order $\{2,1,3\}$ such that each segment was attempted in 10 manual trials and in 10 robotic trials. While these trials represented the first attempts by the operator to perform the overall device delivery task, they benefited from the experience of having developed the system. 

For all experiments, the operator was instructed to deliver the clip using camera image guidance such that it was placed in the middle of the specified leaflet segment with the clip axis parallel to the axis of the valve and with the arms orthogonal to the leaflet line of coaptation. 

All trials consisted of performing steps 2-8. The trial was initialized with the clip extended from the transseptal sheath in a standard orientation inside the left atrium. For the manual trials, the catheter handles were reset to a consistent configuration while, for the robotic system, the motor rotations were reinitialized. Step 4 was also modified in that the clip arms were not opened to aid in orienting them orthogonal to the leaflet coaptation line. While opening the arms is useful to enhance visualization under ultrasound, accurate roll determination did not require this using optical camera imaging. Finally, in step 8, the clip was closed but not released since it is not possible to reload a released clip into the delivery system. 

Task performance was assessed based on trial time as well as clip positioning as described below.

\subsubsection{Procedure time}

For the six sets of trials, it was observed that operator completion time decreased over the first five trials and was consistent over the last five trials. Consequently, timing data for the last five trials was analyzed. To understand how how delivery time was divided between the steps of the procedure, video of the trials was used to extract timings for groups of steps as follows. Steps 2 and 3, representing the initial positioning of the clip above a leaflet segment, formed the first group. Steps 4 through 6 formed the second group. They involved rolling the device sheath so that the clip arms are orthogonal to the leaflet, translating the clip below the leaflets, opening the arms and retracting to capture the leaflets. 

We found that rolling the device sheath inadvertently changed the clip position creating the need to return to step 3 to readjust the clip position. In addition, if the device sheath did not properly release the built-up torque after rolling the clip, translation could inadvertently change the clip orientation, requiring correction. Taken together, the total time to iteratively repeat steps 3-5 is referred to as Step 3' to 5'. Finally, steps 7 and 8, comprising lowering the grippers and closing the clip arms, represent the final timed group.

\begin{figure}[h!]
        \centering
        \includegraphics[width=0.49\textwidth]{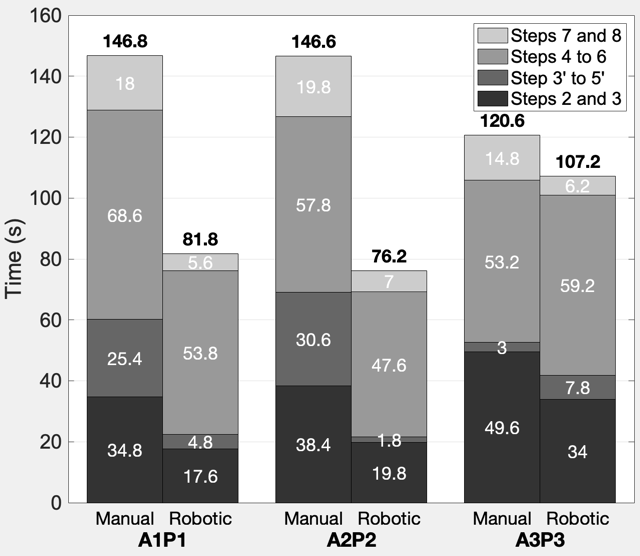}
        \caption{Total delivery time decomposed into procedure steps. Step 3' to 5' is the time repeating these steps to correct for inadvertant catheter motions.}
        \label{time}
        \end{figure}

The results using this breakdown are given in Figure \ref{time}. For each of the three locations, robotic procedure time is less than manual time. For, A2P2, the most frequent clip placement location, the robotic time is 52\% of the manual time. For A1P1 and A2P2, most of the extra time required for the manual procedure is associated with Step 3’ to 5’. This was time spent correcting for inadvertent motions of the clip. In contrast, these problems were largely eliminated by the robotic system.

\subsubsection{Clip Placement Accuracy}
\begin{figure}[h!]
        \centering
        \includegraphics[width=0.45\textwidth]{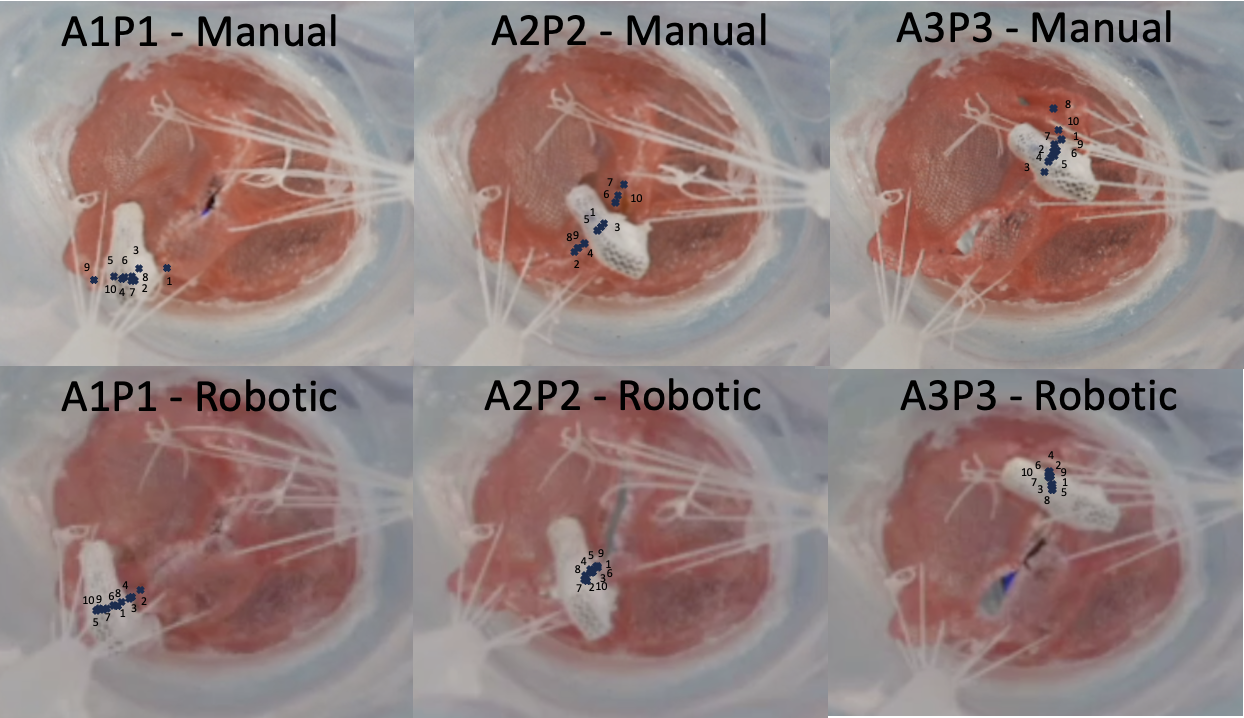}
        \caption{Accuracy of manual versus robotic clip placement.}
        \label{accuracy}
        \end{figure}

To assess the quality of clip placement, final clip location was superimposed for all trials within each of the six sets (Figure \ref{accuracy}). The operator was instructed to place the clip at the midpoint of each segment and, as can be observed from the trial-based numbering of data points, an improvement with increasing trial number is not evident. This is likely because the operator was instructed that positioning accuracy was more important than procedure time. 

For A2P2 and A3P3, robotic delivery results in more accurate clip placement. For A2P2, manual clip placement over the 10 trials spans a 15mm length along the leaftlet free edge. Robotic placement reduces the spread to 4mm. While robotic placement at A1P1 produces a larger spread of points, they all follow the curve of the coaptation line. In contrast, the manually placed clips on this segment are not all located on the coaptation line indicating that the clips were not placed parallel to the axis of the valve.

\section{Conclusion} 

This study demonstrates that robotic assistance can substantially mitigate the challenges of manual transcatheter device delivery. In the context of TEER performed in a phantom model of the heart and vasculature, the use of robotics effectively reduced the inadvertent motion problems identified in Table \ref{tab:limitations}, improved time efficiency and produced more accurate device placements. Although formal cognitive load data was not collected, the operator reported that robotic catheter control was intuitively responsive and less mentally taxing than manual catheter operation, supporting the notion that robotics may lower operator burden during complex navigation tasks.

This work reinforces several important concepts in robotic-assisted catheter navigation. First, the manual system's complex and unintuitive motion control can be substantially improved upon with robotic assistance. While more advanced control mappings are likely to further enhance performance, our findings indicate that even a basic joint-based controller can provide substantial benefit. Crucially, decomposing the procedure into discrete steps and aligning control mappings with standard clinical imaging views proved to be a key strategy. We believe that this stepwise control paradigm can be extended to other transcatheter valve procedures, facilitating the development of a robotic valve repair and replacement platform.

\bibliographystyle{IEEEtran}
\bibliography{references}
\end{document}